\documentclass{llncs}

\usepackage{url}
\usepackage{graphicx}

\usepackage{tikz}
\usetikzlibrary{arrows.meta, backgrounds, calc, positioning, shapes.geometric}

\usepackage{placeins}
\usepackage{relsize}
\usepackage{booktabs}
\usepackage{amsfonts}
\usepackage{amsmath}
\usepackage{mathtools}
\usepackage{wrapfig}
\usepackage{multirow}
\usepackage{bbding}
\usepackage{cleveref}
\usepackage{pdfrender}

\newcommand{\R}{\mathbb{R}}

\newcommand{\MLP}{\text{MLP}}
\newcommand{\ctd}{\text{ConTP}}

\newcommand{\BF}[1]{%
  \textpdfrender{
    TextRenderingMode=FillStroke,
    LineWidth=.2pt, 
  }{#1}%
}\newcommand{\UL}[1]{\underline{#1}}

\newcommand{\model}{IMTS-Mixer}

\definecolor{myred}{RGB}{220,100,120}   
\definecolor{mygreen}{RGB}{174,213,129} 
\definecolor{myblue}{RGB}{100,165,255} 

\definecolor{softred}{RGB}{200, 155, 155}   
\definecolor{softgreen}{RGB}{155, 200, 155} 
\definecolor{softblue}{RGB}{155,155,200} 

\definecolor{verysoftred}{RGB}{197, 202, 233}

\definecolor{relublue}{RGB}{173, 216, 230}  
\definecolor{verylightblue}{RGB}{220,240,255} 
\definecolor{pastelyellow}{RGB}{255, 245, 179}  

\definecolor{lightgreen}{RGB}{149, 203, 160}
\definecolor{modelcolor}{RGB}{253, 233, 196} 
\definecolor{transgray}{RGB}{224, 224, 224}

\definecolor{verylightgreen}{RGB}{220,255,220} 

\title{Mixing It Up: Exploring Mixer Networks for Irregular Multivariate Time Series Forecasting	}
\author{Christian Klötergens\inst{1,2} \Envelope \and
Tim Dernedde \inst{1} \and
Lars Schmidt-Thieme\inst{1,2} \and 
Vijaya Krishna Yalavarthi\inst{1,2}
}

\authorrunning{C. Klötergens et al.}

\institute{Institute of Computer Science, \ University of Hildesheim, Hildesheim, Germany \\
\email{\{kloetergens, dernedde, schmidt-thieme,yalavarthi\}@ismll.de}
\and
VWFS Data Analytics Research Center (VWFS-DARC), Hildesheim, Germany}
\begin{document}
\maketitle


\begin{abstract}	
Forecasting irregularly sampled multivariate time series with missing values (IMTS) is a fundamental challenge in domains such as healthcare, climate science, and biology. 
While recent advances in vision and time series forecasting have shown that lightweight MLP-based architectures (e.g., MLP-Mixer, TSMixer) can rival attention-based models in both accuracy and efficiency, their applicability to irregular and sparse time series remains unexplored. 
In this paper, we propose \textbf{\model{}}, a novel architecture that adapts the principles of Mixer models to the IMTS setting. 
\model{} introduces two key components: (1) \textbf{ISCAM}, a channel-wise encoder that transforms irregular observations into fixed-size vectors using simple MLPs, and (2) \textbf{\ctd{}}, a continuous time decoder that supports forecasting at arbitrary time points. 
In our experiments on established benchmark datasets we show that our model achieves state-of-the-art performance in both forecasting accuracy and inference time, while using fewer parameters compared to baselines.
\end{abstract}


\section{Introduction}
\begin{wrapfigure}{r}{0.5\textwidth}
    \centering
    \begin{tikzpicture}[x=0.7cm, y=0.7cm]


        \normalsize
        \draw[-latex] (0,0.5) -- (7.5,0.5);
        \node at (7.15,0.5)[below=0cm]{\scriptsize Time};

        \draw[very thick, myblue, smooth] plot coordinates {
            (1, 1) (1.66, 1.5) (2.32, 1.2) (2.98, 1.4) (3.64, 0.8)
            (4.3, 1.1) (4.96, 0.7) (5.62, 0.7) (6.28, 0.9) (6.94, 0.6)
        };

        \foreach \x in  {1,1.3,3.64,2.98,3.64,4,4.96,2.32,5.62,5.9,6.94,6.28}{
            \draw (\x,0.42) -- (\x,0.58); 
        }

        \draw[very thick, myred, smooth] plot coordinates {
            (1, 1.5) (1.66, 0.7) (2.32, 0.7) (2.98, 1) (3.64, 1.1)
            (4.3, 1.4) (4.96, 1.2) (5.62, 0.9) (6.28, 0.8) (6.94, 1.2)
        };

        \draw[very thick, mygreen, smooth] plot coordinates {
            (1, 2) (1.66, 1.8) (2.32, 1.6) (2.98, 1.9) (3.64, 2.1)
            (4.3, 2.3) (4.96, 2) (5.62, 1.8) (6.28, 1.6) (6.94, 1.5)
        };

        \foreach \x/\y in {
            1.3/1.07, 2.32/0.7,  4.96/1.2} {
                \draw[fill=myred] (\x,\y) circle (0.07cm);
        }
        \foreach \x/\y in {6.28/0.8} {
                \draw[fill=myred, opacity=0.5] (\x,\y) circle (0.07cm);
        }

        \foreach \x/\y in {
        2.98/1.4, 3.64/0.8, 4.0/0.93} {
                \draw[fill=myblue] (\x,\y) circle (0.07cm);
        }

        \foreach \x/\y in {5.62/0.7} {
                \draw[fill=myblue,opacity=0.5] (\x,\y) circle (0.07cm);
        }

        \foreach \x/\y in {
            1/2, 3.64/2.1} {
                \draw[fill=mygreen] (\x,\y) circle (0.07cm);
        }

        \foreach \x/\y in {
            5.9/1.72, 5.62/1.8, 6.94/1.5} {
                \draw[fill=mygreen, opacity=0.5] (\x,\y) circle (0.07cm);
        }

        \draw[dashed] (5.25,0.5) -- (5.25,2.8) ; 

        \node[left] at (1,2) {$\text{c}_1$};
        \node[left] at (1,1.5) {$\text{c}_2$};
        \node[left] at (1,1) {$\text{c}_3$};

        \scriptsize
        \node[] at (2.625,2.6) {Observation};
        \node[] at (2.625,2.3) {Range};
        \node[] at (6.35,2.6) {Forecasting};
        \node[] at (6.35,2.3) {Horizon};
        \normalsize

    \end{tikzpicture}
    \caption{Example of an IMTS Forecasting task. The observations and forecasting targets are irregularly spaced.}\label{fig:time_series_curvy}
\end{wrapfigure}
Time series forecasting is a critical task across various domains.
While a significant body of research has focused on forecasting regularly sampled, fully observed multivariate time series, many real-world applications, such as healthcare, economics, astronomy, and climate, deal with time series data that are \emph{irregularly} sampled, with observations missing across different channels and timestamps.
When such data are aligned to a global timeline, the result is a highly sparse and uneven multivariate structure, as illustrated in~\Cref{fig:time_series_curvy}. 
We refer to such data as \textbf{irregularly sampled multivariate time series with missing values (IMTS)}.
Traditional approaches to modeling IMTS have relied on Neural-ODE frameworks, where the time series is modeled as the solution to an ordinary differential equation (ODE) with a neural network-defined vector field~\cite{Rubanova2019.Latent,DeBrouwer2019.GRUODEBayes,Schirmer2022.Modeling}.
While powerful in principle, these models tend to be computationally expensive due to the sequential nature of ODE solvers, and they lack a systematic way to handle missing values.
Recently, attention-based models have emerged as a more efficient and accurate alternative, achieving competitive performance on IMTS forecasting tasks~\cite{Yalavarthi2023.Forecasting,Zhang.Irregular,Liu2025.Timecheat}.
These models typically transform irregular channel-wise observations into fixed-size representations and apply attention mechanisms to capture inter- and intra-channel dependencies.
However, such models are often resource-intensive, requiring a large number of parameters and high memory \mbox{requirements}.

A recent work in computer vision has demonstrated that carefully arranged MLP-based models can rival or even surpass attention-based models~\cite{Tolstikhin2021.MLPMixer}.
This insight has also been applied to forecasting tasks with regularly sampled time series~\cite{Ekambaram2023.TSMixer,Chen2023.TSMixer}.
However, this line of research has not yet been extended to irregularly sampled time series with missing values.
In this paper, we introduce \textbf{\model{}}, a novel architecture for forecasting with IMTS that combines the simplicity of MLP-based designs with specialized modules to address irregular sampling and forecasting at arbitrary time points.
First, we propose the \textbf{I}rregularly \textbf{S}ampled \textbf{C}hannel \textbf{A}ggregation \textbf{M}odule (ISCAM), a lightweight encoder that maps each irregular channel into a fixed-size representation using simple MLPs. These embeddings are then concatenated and processed by an MLP-Mixer-style architecture to learn inter- and intra-channel interactions.
Second, to support continuous-time forecasting, we introduce the \textbf{Con}tinuous \textbf{T}emporal \textbf{P}rojection (\textbf{\ctd}) module, a decoder that takes the mixer output, a continuous-valued target timestamp, and a channel identifier to generate a prediction.
In an extensive comparison, we evaluate \model{} on four real-world datasets from various domains.
We observe that \model{} establishes a new state-of-the-art in terms of forecasting accuracy on three out of four datasets. 
Our contributions are summarized as follows\footnote{The code is available here: \url{https://github.com/kloetergensc/IMTS-Mixer/}}:
\vspace{-0.5cm}
\begin{enumerate}
	\itemsep0cm 
	\item We propose \textbf{\model{}}, a novel MLP-based architecture tailored for irregularly sampled multivariate time series with missing values (IMTS), combining efficient channel aggregation and continuous-time prediction.
	
	\item We introduce \textbf{ISCAM}, a simple and effective channel-wise encoder that maps irregularly observed channels to fixed-size representations, enabling MLP processing.
	
	\item We develop a lightweight temporal projection module (\textbf{\ctd}) that maps a fixed-size representation of an input channel to future IMTS predictions.
\end{enumerate}

\section{Related Work}


\subsection{Regular Multivariate Forecasting}
The Regular Multivariate Time Series forecasting literature recently focused on transformer-based~\cite{Vaswani2017.Attention} architectures~\cite{Zhou2021.Informer,Wu2021.Autoformer,Zhou2022.FEDformer,Nie2022.Time,Liu2023.ITransformer}. 
However, it was shown that a simple model based on linear layers (DLinear~\cite{Zeng2023.Are}) can be competitive and partially outperform some of the more complex transformer architectures. 
Additionally, mixer architectures have been adapted from the computer vision domain~\cite{Tolstikhin2021.MLPMixer} to time series~\cite{Ekambaram2023.TSMixer,Chen2023.TSMixer}.
These models exclusively rely on fully-connected networks that are alternately applied along the channel and time dimension and achieve state-of-the-art results. 

\subsection{IMTS Forecasting}
Initially, researchers relied on Neural ODE-based~\cite{Chen2018.Neural} approaches to model irregular time series~\cite{Rubanova2019.Latent,DeBrouwer2019.GRUODEBayes,Bilos2021.Neural,Schirmer2022.Modeling}. 
ODEs are the prevalent framework in science and engineering to model how systems evolve over time in a continuous manner. 
However, using neural networks to represent ODE systems has proven to be less effective than other IMTS modeling approaches and also incurs relatively long inference times~\cite{Yalavarthi2023.Forecasting,Zhang.Irregular}. 

\paragraph{Graph Transformers for IMTS forecasting.}
Apart from the ODE-based approaches the Transformable Patch Graph Neural Network (tPatchGNN)~\cite{Zhang.Irregular} applies a patching mechanism to each channel. 
The patches are then processed with a Transformer and serve as input for a Graph Neural Network (GNN) that models inter-channel dependencies. 
Additionally, we want to highlight Graphs for Forecasting Irregular Multivariate Time Series (GraFITi)~\cite{Yalavarthi2023.Forecasting}. 
Here, the IMTS is represented with a graph in which channels and time steps are nodes and observed values are stored in the edges between the corresponding time and channel nodes. 
Forecasts are made by predicting the edge values of query-time nodes with graph attention layers. 
While this approach provides state-of-the-art performance, it also comes with major drawbacks that relate to the graph structure itself. 
GraFITi relies on time nodes that are connected to multiple channel nodes, corresponding to observation steps that observe more than one channel. 
If only one channel is observed at a time, the resulting graph is disconnected, rendering GraFITi incapable of modeling inter-channel interactions. 
Another problem of GraFITi lies in the fact that the query nodes are indirectly connected and hence influence each other. 
This entails that predictions will change depending on what \emph{other} queries we give to the model. 

TimeCHEAT~\cite{Liu2025.Timecheat} transforms an IMTS into regularly observed patches using a GraFITi-based 
encoder. These patches are then further processed by a transformer and a linear decoder to obtain the final predictions.

\section{Mixer Networks}\label{sec:prelims_mixer}

\paragraph{MLP-Mixer.} 
The MLP-Mixer~\cite{Tolstikhin2021.MLPMixer} was originally proposed as a simple and efficient alternative to convolutional neural networks (CNNs) and Vision Transformers (ViTs) for image classification tasks. In this architecture, a 2D image ($\in \R^{W\times H}$, where $W,H$ are width and height) is divided into $P$ non-overlapping patches, which are flattened and projected into vectors of length $C$, resulting in an input representation $X \in \mathbb{R}^{P \times C}$.

The core of the MLP-Mixer consists of a stack of mixer layers, each containing two separate MLP blocks:
\begin{itemize}
	\item a \textbf{token-mixing MLP} $\MLP_P: \mathbb{R}^P \rightarrow \mathbb{R}^P$ operating across patches (rows of $X$),
	\item a \textbf{channel-mixing MLP} $\MLP_C: \mathbb{R}^C \rightarrow \mathbb{R}^C$ operating across channels (columns of $X$).
\end{itemize}
Each MLP block is preceded by layer normalization and followed by a residual connection. Formally, given input $X \in \mathbb{R}^{P \times C}$, the output of a single mixer layer is computed as:
\begin{align*}
	U_{*,c} & = X_{*,c} + \text{LayerNorm}\left(\MLP_C(X_{*,c})\right), \; \text{ } c = 1, \ldots, C \\
	Z_{p,*} & = U_{p,*} + \text{LayerNorm}\left(\MLP_P(U_{p,*})\right), \; \text{ } p = 1, \ldots, P
\end{align*}
Here, $X_{*,c}$ denotes the $c$-th column (channel vector), and $X_{p,*}$ denotes the $p$-th row (patch vector).

Despite its simplicity and lack of attention mechanisms, MLP-Mixer has demonstrated competitive performance, often outperforming more complex models such as Vision Transformers~\cite{Dosovitskiy2020.Image} on various image classification benchmarks.

\paragraph{TSMixer.}
Motivated by the success of MLP-Mixer in vision, \cite{Chen2023.TSMixer} proposed \emph{TSMixer}, an adaptation of the Mixer architecture for regular multivariate time series forecasting. The goal is to predict the next $Q$ time steps $Y \in \mathbb{R}^{Q \times C}$ from a past window of $T$ time steps $X \in \mathbb{R}^{T \times C}$, where $C$ denotes the number of variables (channels).
TSMixer applies a stack of Mixer layers to the input $X$, where:
\begin{itemize}
	\item the \textbf{temporal mixer} (analogous to token-mixing in images) mixes information across time steps: $\MLP_T: \mathbb{R}^T \rightarrow \mathbb{R}^T$,
	\item the \textbf{channel mixer} operates across variables: $\MLP_C: \mathbb{R}^C \rightarrow \mathbb{R}^C$.
\end{itemize}
Similar to MLP-Mixer, these mixers are applied alternately with residual connections and layer normalization. After the stack of mixer layers, a final linear projection $\mathbb{R}^{T} \rightarrow \mathbb{R}^{Q}$ is applied across the temporal axis to produce the forecast.

\section{The IMTS Forecasting Problem}

We define an \emph{irregularly sampled multivariate time series (IMTS)} as a collection of $C$ unaligned, sparse, and independently sampled univariate time series, one per channel:
$
X \coloneqq [X_1, \ldots, X_C] \in \mathcal{X},
$
where each channel $X_c \in \textnormal{Seq}(\mathbb{R} \times \mathbb{R})$ is a sequence of observation tuples
$
X_c \coloneqq \left(\left(t_{c,i}, v_{c,i}\right)\right)_{i=1}^{N_c}.
$
Here, $t_{c,i} \in \mathbb{R}$ refers to the timestamp of the $i$-th observation in channel $c$
and $v_{c,i} \in \mathbb{R}$ is the respective observed value.
Channels may be observed at different timestamps and with different sequence lengths $N_c$, resulting in irregular and sparse structure when aligned over a global timeline.

\paragraph{Forecasting Query.}
Analogous to the IMTS definition, we define a \emph{forecasting query} as a collection of per-channel timepoints at which future values should be predicted:
$
Q \coloneqq [Q_1, \ldots, Q_C] \in \mathcal{Q}, $ where $
Q_c \left(q_{c,i}\right)_{i=1}^{K_c} \in \textnormal{Seq}(\mathbb{R}).
$
Each $q_{c,i}$ specifies a future timestamp for which a prediction is required for channel $c$. 
The corresponding \textbf{forecasting answer} is given by:
$Y = [Y_1, \ldots, Y_C] \in \mathcal{Y}$, where $Y_c = \left(y_{c,i}\right)_{i=1}^{K_c} \in \textnormal{Seq}(\mathbb{R}),$
ensuring $|Q_c| = |Y_c| = K_c$ for each channel $c$.

\begin{problem}\label{def:prob}
	\textbf{IMTS Forecasting Problem.}
	Given a training dataset of $M$ triples:
	\(
	\{(X^{(m)}, Q^{(m)}, Y^{(m)})\}_{m=1}^{M}
	\)
	drawn from an unknown distribution $\rho$ over $\mathcal{X} \times \mathcal{Q} \times \mathcal{Y}$, and a loss function $\ell: \mathcal{Y} \times \mathcal{Y} \rightarrow \mathbb{R}$ (e.g., mean squared error), the goal is to learn a model:
	\(
	\hat{Y}: \mathcal{X} \times \mathcal{Q} \rightarrow \mathcal{Y}
	\)
	that minimizes the expected loss:
	\begin{align}
		\mathcal{L}(\hat{Y}; \rho) := \mathbb{E}_{(X, Q, Y) \sim \rho} \left[ \ell(Y, \hat{Y}(X, Q)) \right]
	\end{align}
\end{problem}

\section{IMTS-Mixer}

Each IMTS consists of $C$ channels, where the number of observations $N_c$ for each channel $c$ may vary, both within a single IMTS and across different samples. This irregularity makes it challenging to directly apply standard neural architectures, particularly MLP-Mixers, which require fixed-size input representations.

To address this, we propose the \textbf{Irregularly Sampled Channel Aggregation Module (ISCAM)}, which encodes each variable-length channel $X_c$ into a fixed-dimensional vector $Z_c \in \mathbb{R}^D$, enabling uniform downstream processing using MLP-Mixer blocks.

\subsection{Irregularly Sampled Channel Aggregation Module}\label{sec:iscam}

Let a single channel $X_c$ represent a sequence of $N_c$ observations. Our goal is to transform this irregular sequence into a fixed-size embedding $Z_c \in \mathbb{R}^D$.
For clarity, we drop the channel subscript $c$ and denote the observations as $(t_i, v_i)$.

\paragraph{Observation-Tuple Embedding.}
Each observation tuple $(t_i, v_i)$ is embedded using a shared MLP $f_\textsc{ote}: \mathbb{R}^2 \to \mathbb{R}^D$, resulting in:
\[
H = [h_1, \ldots, h_{N_c}] \in \mathbb{R}^{N_c \times D}, \quad \text{where } h_i = f_\textsc{ote}([v_i, t_i])
\]

\paragraph{Weighted Aggregation.}
To compute an importance score for each observation, we apply a second shared MLP $f_\textsc{wa}: \mathbb{R}^2 \to \mathbb{R}^D$:
\[
A = [a_1, \ldots, a_{N_c}] \in \mathbb{R}^{N_c \times D}, \quad \text{where } a_i = f_\textsc{wa}([v_i, t_i])
\]

Unlike standard attention mechanisms which compute a scalar weight per observation, we assign a separate weight to each embedding feature, similar to multi-head attention, but without cross-token interactions.

To handle the variable length $N_c$ across samples and ensure scale invariance, we apply softmax normalization along the columns $d$:
\[
\text{softmax}: \mathbb{R}^{N_c} \to \mathbb{R}^{N_c}, \quad \text{so that } \sum_{i=1}^{N_c} \text{softmax}(a_{i,d}) = 1
\]
Let $A_{:,d} = (a_{1,d}, \ldots, a_{N_c,d})$ and $H_{:,d} = (h_{1,d}, \ldots, h_{N_c,d})$ denote the $d$-th column across the $N_c$ elements of $A$ and $H$, respectively. The final channel embedding is then computed as:
\begin{align}\label{eq:agg}
	Z_c = \left[ \sum_{i=1}^{N_c} \text{softmax}(A_{:,d})_i \cdot H_{:,d} \right]_{d=1}^D \in \mathbb{R}^D
\end{align}
This yields a fixed-size vector per channel, where each dimension is an importance-weighted average over time.
While one can have channel-specific parameters in $f_\textsc{ote}$ and $f_\textsc{wa}$ we allow to share these weights among channels.

\paragraph{Channel Bias.}
Since the embedding MLPs are shared across all channels, global channel-specific information may be lost. To address this, we introduce a \emph{channel bias} vector $b_c \in \mathbb{R}^D$ for each channel $c$, composed of learnable parameters. The final channel encoding is computed as:
$Z_c^+ = Z_c + b_c \in \mathbb{R}^D$.
This allows the model to retain per-channel identity and compensate for missing patterns or context. 
In the case where a channel $c$ is missing (i.e., $X_c$ is empty), we define $Z_c := 0$ so that the output $Z_c^+ = b_c$ contains only the bias information.

\paragraph{Comparison with Existing Architectures.}
Several prior models such as SeFT~\cite{Horn2020.Set}, Tripletformer~\cite{Yalavarthi2023.Tripletformer}, and GraFITi~\cite{Yalavarthi2023.Forecasting} use attention-based mechanisms over IMTS observations, where the aggregation is performed via a softmax-weighted sum of encoded tuples. These methods compute weights through the scaled dot-product of learned queries and keys, following the standard attention framework~\cite{Vaswani2017.Attention}.

In contrast, ISCAM bypasses query-key interactions entirely by modeling importance weights directly through a learned MLP applied to each observation tuple. This results in a simpler and more lightweight design, while still enabling dimension-specific weighting through softmax normalization.

TTCN~\cite{Zhang.Irregular,Zhang2024.Irregulara} also employs MLP-based aggregation over irregular time series. However, it introduces a more complex mechanism by using separate MLPs (called filters) for each embedding dimension $d$. Moreover, TTCN lacks a dedicated channel bias component, which limits its ability to encode global per-channel information independent of input observations.


\subsection{Mixer Blocks}
Similar to MLP-Mixer and TSMixer, we employ $L$ mixer blocks, each operating over the channel and feature dimensions. A single mixer block at layer $l$ updates the representation $Z^{(l-1)} \in \mathbb{R}^{C \times D}$ as follows:

\begin{align}
	\begin{aligned}
		Z^{\prime{(l)}} &= Z^{(l-1)} + \text{ReLU} \left( \text{Linear}_\textsc{chan}^{(l)} \left( \text{RMS} ( {Z^{(l-1)}}^\top) \right) \right)^\top \\
		Z^{(l)} &=  Z^{(l-1)} + Z^{\prime{(l)}} + \text{ReLU} \left( \text{Linear}_\textsc{dim}^{(l)} \left( \text{RMS} (Z^{\prime{(l)}} ) \right) \right)
	\end{aligned}
\end{align}

Here,
$\text{Linear}_\textsc{chan}^{(l)}$ operates along the channel dimension (i.e., across $C$ channels),
$\text{Linear}_\textsc{dim}^{(l)}$ operates along the feature dimension ($D$),
RMS denotes RMSNorm~\cite{Zhang2019.Root}, and
ReLU is the element-wise nonlinearity.

Compared to MLP-Mixer and TSMixer described in Section~\ref{sec:prelims_mixer}, we introduce two modifications to adapt to the irregular multivariate forecasting setting:
\begin{enumerate}
	\item \textbf{Flexible Output Dimension:} We allow the output of the final mixer layer to have a configurable feature dimension $D_\textsc{out}$, i.e., $Z^{(L)} \in \mathbb{R}^{C \times D_\textsc{out}}$. This makes it possible to match the projection to varying numbers of queries per channel. All intermediate layers maintain a fixed hidden size $D$.
	
	\item \textbf{RMSNorm Instead of LayerNorm:} We replace LayerNorm with RMSNorm~\cite{Zhang2019.Root}, which has been shown to yield faster convergence while maintaining predictive performance.
\end{enumerate}

\subsection{Continuous Temporal Projection}
TSMixer utilizes a linear layer to project the encoded representation of each channel to forecasts on a predefined regular grid of query steps~\cite{Chen2023.TSMixer}. 
This design, however, is incompatible with IMTS, where query times are not fixed but vary continuously across instances.

To address this, \model{} introduces a \textbf{Continuous Temporal Projection} (\textbf{\ctd{}}), which allows the model to project the hidden state of each channel to any given point in time.
Instead of learning linear layers tied to predetermined query steps, \ctd{} learns a function $f_\textsc{qu}: \mathbb{R} \rightarrow \mathbb{R}^{D_\textsc{out}}$ that maps scalar query times to the weights of a linear projection dynamically. 
Formally, the forecast for channel $c$ at query time $q_{c,i}$ is computed as:
\begin{align}
	\begin{aligned}
		\hat{y}_{i,c} &= f_\textsc{qu}^c(q_{c,i}) \cdot Z_c^{(L)} + b^{\textsc{out}}_c \in \mathbb{R}
	\end{aligned}
\end{align}
In order to enhance expressiveness, we learn a separate function $f_\textsc{qu}^c$ for each channel and implement this with a two-layer MLP.\@
Additionally, we learn a query independent output bias $b^\textsc{out} \in \R^C$.
The complete architecture of \model{} is illustrated in Figure~\ref{fig:model}.
\newsavebox{\mixer}
\savebox{\mixer}{
    \begin{tikzpicture}
        
    \node at (2.5, 0.0){{${Z^{l-1}}$}};

    \draw[-latex, thick] (2.5,0.25) to (2.5,0.65);         
    \draw[ thick,rounded corners, fill=transgray] (1.5,0.65) rectangle (3.5,1.15);
    \draw[-latex, thick] (2.5,1.5) to (2.5,1.85); 
    \node at (2.5, 0.9){\textsf{transpose}};
    \draw[-latex,thick] (2.5,1.15) to (2.5,1.5);
    \draw [rounded corners=1pt, thick, fill=transgray] (1,1.5) rectangle (4,3.0);
    \node at (2.5, 1.7){\small{\textsf{RMS-Norm}}};
    \draw [fill=verysoftred,rounded corners=0pt, thick](1,1.9) rectangle (4,2.6);
    \node at (2.5, 2.25){\textsf{Fully-Connected}};
    \node at (2.5, 2.8){\small{\textsf{ReLU}}};
    \draw[-latex, thick] (2.5,3.0) to (2.5,3.3);

    \draw[rounded corners,  thick, fill=transgray] (1.5,3.3) rectangle (3.5,3.8);
    \node at (2.5, 3.55){\textsf{transpose}};
    \draw[-latex,  thick] (2.5,3.8) to (2.5,4.3);

    \draw[ thick, rounded corners=1pt, fill=transgray,] (1,4.3) rectangle (4,5.8);
    \node at (2.5, 4.5){\small{\textsf{RMS-Norm}}};
    \draw[fill=verysoftred,rounded corners=0pt, thick] (1,4.7) rectangle (4,5.4);
    \node at (2.5, 5.05){\textsf{Fully-Connected}};
    \node at (2.5, 5.6){\small{\textsf{ReLU}}};

    \draw[-latex,  thick] (2.5,5.8) to (2.5,6.15);
    \node at (2.5, 6.45){{${Z^{l}}$}};

    \draw[-latex,  thick, rounded corners] (3,0) -- (4.7,0) -- (4.7,4.0) -- (2.5,4.0);
    \draw[-latex,  thick, rounded corners] (2,0) -- (0.5,0) -- (0.5,6.45) -- (2,6.45);
    \draw[-latex,  thick, rounded corners] (3.5,3.55) -- (4.5,3.55) -- (4.5,6.45) -- (3,6.45);

    \node[right] at (0.5,3.225) {\large{\textbf{+}}};
    \node[left] at (4.7,2) {\large{\textbf{+}}};
    \node[left] at (4.55,5) {\large{\textbf{+}}};
    \end{tikzpicture}
}

\newsavebox{\zmatrix}
\savebox{\zmatrix}{
    \begin{tikzpicture}[scale=0.5]
        \foreach \j in {0,1,2,3} {
            \draw[fill=softgreen,  opacity=0.9] (\j,0) rectangle ++(1,-1);
        }
        \foreach \j in {0,1,2,3} {
            \draw[fill=softred, opacity=0.9] (\j,-1) rectangle ++(1,-1);
        }
        \foreach \j in {0,1,2,3} {
            \draw[fill=softblue, opacity=0.9] (\j,-2) rectangle ++(1,-1);
        }
    \end{tikzpicture}
}

\newsavebox{\zvectorgreen}
\savebox{\zvectorgreen}{
    \begin{tikzpicture}[scale=0.5]
        \foreach \j in {0,1,2,3} {
            \draw[fill=softgreen, opacity=0.9] (\j,0) rectangle ++(1,-1);
        }
    \end{tikzpicture}
}

\newsavebox{\zvectorblue}
\savebox{\zvectorblue}{
    \begin{tikzpicture}[scale=0.5]
        \foreach \j in {0,1,2,3} {
            \draw[fill=softblue,  opacity=0.9] (\j,0) rectangle ++(1,-1);
        }
    \end{tikzpicture}
}

\newsavebox{\zvectorred}
\savebox{\zvectorred}{
    \begin{tikzpicture}[scale=0.5]
        \foreach \j in {0,1,2,3} {
            \draw[fill=softred,  opacity=0.9] (\j,0) rectangle ++(1,-1);
        }
    \end{tikzpicture}
}

\begin{figure*}
    \centering
    \resizebox{\textwidth}{!}{
    \begin{tikzpicture}[scale=0.97]


        \node[rotate=90] at (0,0.5) {\scriptsize{\textsf{Channel $\mathsf{C}$}}};
        \node[rotate=270] at (0,1.5) {\textsf{$\ldots$}};
        \node[rotate=90] at (0,2.5) {\scriptsize{\textsf{Channel $\mathsf{2}$}}};
        \node[rotate=90] at (0,4.5) {\scriptsize{\textsf{Channel $\mathsf{1}$}}};


        \draw[black, thick] (0.25, 0) rectangle (3.75, 1);
        \draw[softblue, smooth, thick] plot coordinates {
            (0.5, 0.5) (0.83, 0.75) (1.16, 0.6) (1.49, 0.7) (1.82, 0.4)
            (2.15, 0.55) (2.48, 0.35) (2.81, 0.35) (3.14, 0.45) (3.47, 0.3)
        };
        \foreach \x/\y in {1.49/0.7, 1.82/0.4, 2.0/0.465} {
            \draw[fill=softblue] (\x,\y) circle (0.07cm);
        }
        \foreach \x/\y in {2.81/0.35} {
            \draw[fill=softblue,opacity=0.5] (\x,\y) circle (0.07cm);
        }
        \draw[  dotted] (2.7,0) -- (2.7,1) ;
        \draw[fill=transgray,opacity=0.3] (2.7,0) rectangle (3.75,1) ; 
        \node at (3.55,0.85) {\scriptsize $\mathsf{Q_C}$}; 

        \draw[](0.25,1) -- (0.25,1.1) -- (2.7,1.1) -- (2.7,1);  
        \draw[-latex, rounded corners, thick](1.475,1.145) -- (1.475,1.45) -- (4.0,1.45) --(4.0,3) -- (4.5,3) ;
        \draw[fill=modelcolor, thick] (2.2, 1.3) rectangle (3.4,1.6);
        \node at(2.8,1.45) {\scriptsize{\textsf{ISCAM}}};
    
        \draw[black, thick] (0.25, 2) rectangle (3.75, 3);
        \draw[softred, smooth, thick] plot coordinates {
            (0.5, 2.75) (0.83, 2.35) (1.16, 2.35) (1.49, 2.5) (1.82, 2.55)
            (2.15, 2.7) (2.48, 2.6) (2.81, 2.45) (3.14, 2.4) (3.47, 2.6)
        };
        \foreach \x/\y in {0.65/2.535, 1.16/2.35, 2.48/2.6} {
            \draw[fill=softred] (\x,\y) circle (0.07cm);
        }
        \foreach \x/\y in {3.14/2.4} {
            \draw[fill=softred, opacity=0.5] (\x,\y) circle (0.07cm);
        }
        \draw[dotted] (2.7,2) -- (2.7,3) ;
        \draw[fill=transgray,opacity=0.3] (2.7,2) rectangle (3.75,3) ; 
        \node at (3.55,2.85) {\scriptsize $\mathsf{Q_2}$}; 

        \draw[](0.25,3) -- (0.25,3.1) -- (2.7,3.1) -- (2.7,3);  
        \draw[-latex,rounded corners, thick](1.475,3.1) -- (1.475,3.45) -- (4.0,3.45) --(4.0,3.75) -- (4.5,3.75) ;
        \draw[fill=modelcolor, thick] (2.2, 3.3) rectangle (3.4,3.6);
        \node at(2.8,3.45) {\scriptsize{\textsf{ISCAM}}};

        \draw[black, thick] (0.25, 4) rectangle (3.75, 5);
        \draw[softgreen, smooth, thick] plot coordinates {
                (0.5, 4.5) (0.83, 4.4) (1.16, 4.3) (1.49, 4.45) (1.82, 4.55)
                (2.15, 4.65) (2.48, 4.5) (2.81, 4.4) (3.14, 4.3) (3.47, 4.25)
        };
        \foreach \x/\y in {0.5/4.5, 1.82/4.55} {
                \draw[fill=softgreen] (\x,\y) circle (0.07cm);
        }
        \foreach \x/\y in {2.95/4.36, 2.81/4.4, 3.47/4.25} {
                \draw[fill=softgreen, opacity=0.5] (\x,\y) circle (0.07cm);
        }
        \draw[dotted,rounded corners=1pt, ] (2.7,4) -- (2.7,5) ;
        \draw[fill=transgray,opacity=0.3] (2.7,4) rectangle (3.75,5) ; 
        \node at (3.55,4.85) {\scriptsize $\mathsf{Q_1}$};

        \draw[](0.25,5) -- (0.25,5.1) -- (2.7,5.1) -- (2.7,5);  
        \draw[-latex, rounded corners, thick](1.475,5.1) -- (1.475,5.45) -- (4.0,5.45) --(4.0,4.5) -- (4.5,4.5) ;
        \draw[fill=modelcolor, thick] (2.2, 5.3) rectangle (3.4,5.6);
        \node at(2.8,5.45) {\scriptsize{\textsf{ISCAM}}};
        
        \node(z1) at  (5.5,4.5) {\scalebox{0.9}{\usebox{\zvectorgreen}}};
        \node(z2) at  (5.5,3.75) {\scalebox{0.9}{\usebox{\zvectorred}}};
        \node(z3) at  (5.5,3) {\scalebox{0.9}{\usebox{\zvectorblue}}};

        \node at (z1) [right=0.9cm] {\scriptsize $Z_1$};
        \node at (z2) [right=0.9cm] {\scriptsize $Z_2$};
        \node at (z3) [right=0.9cm] {\scriptsize $Z_C$};

        \draw[dotted, thick] (4.5,2.5) rectangle (6.5,5);
        \draw[-latex, thick] (5.5, 2.5) -- (5.5, 2.1);
        
        \draw[thick, rounded corners, fill=transgray] (4.75,1.6) rectangle (6.25,2.1) ;
        \node at (5.5,1.85) {\scriptsize \textsf{concatenate}}; 
        \draw[-latex, thick] (5.5, 1.6) -- (5.5, 1.2);

        \node (z) at (5.5,0.5) {\scalebox{0.9}{\usebox{\zmatrix}}};
        \node at (z) [below=0.7cm] {\scriptsize $D$};
        \node at (z) [right=0.9cm] {\scriptsize $C$};

        \draw[-latex, very thick, rounded corners] (5.95,-0.193) -- (5.95,-0.8) -- (9.05,-0.8) -- (9.05,-0.3);

        \draw[fill=modelcolor, opacity=1,rounded corners=4pt, opacity=0.5] (7.4, -0) rectangle (11.3, 5.8) ; 
        \draw[fill=modelcolor,opacity=1,rounded corners=4pt,opacity=0.75] (7.25, -0.15) rectangle (11.15, 5.65) ; 
        \draw[fill=modelcolor, opacity=1,rounded corners=4pt, very thick] (7.1, -0.3) rectangle (11, 5.5) ;
        \node at (9, 2.4) {\scalebox{0.75}{\usebox{\mixer}}};
        \node[above=-1mm] at (7.0,5.55) {\small $\times {L}$} ; 
        \node[right,below] at (10, 5.5) {\footnotesize \textsf{{Mixer Block}}} ;

        \draw[-latex, very thick, rounded corners] (9.35,5.8) -- (9.35,6.2) -- (12.75,6.2) -- (12.75,5);

        \node (zout) at (12.75,4.3) {\scalebox{0.9}{\usebox{\zmatrix}}};
        \node at (zout) [below=0.7cm] {\scriptsize $D_\textsc{out}$};
        \node at (zout) [left=0.9cm] {\scriptsize $C$};

        \draw[-latex, very thick] (12.75,3) -- (12.75,2.5); 

        \node(zout1) at  (12.75,2) {\scalebox{0.9}{\usebox{\zvectorgreen}}};
        \node(zout2) at  (12.75,1.25) {\scalebox{0.9}{\usebox{\zvectorred}}};
        \node(zout3) at  (12.75,0.5) {\scalebox{0.9}{\usebox{\zvectorblue}}};

        \draw[-latex , thick, rounded corners] (13.68,2) -- (13.9,2) --  (13.9,4.5) -- (14.35,4.5) ;
        \draw[-latex , thick, rounded corners] (13.68,1.25) -- (14.05,1.25) --  (14.05,2.5) -- (14.35,2.5) ;
        \draw[-latex , thick, rounded corners] (13.68,0.5) -- (14.35,0.5) ;


        \draw[  fill=modelcolor,opacity=1, rounded corners=1pt, thick] (14.35,0.25) rectangle (15.25,0.75); 
        \node at (14.8,0.5) {\scriptsize \textsf{\ctd}} ;
        \draw[   -latex, thick] (15.25,0.5) -- (15.75,0.5);
        \draw[   fill=transgray,opacity=0.4, rounded corners=1pt, thick] (14.5,-0.6) rectangle (15.1,-0.1);
        \draw[  -latex, very thick, rounded corners=1pt] (14.8,-0.1) -- (14.8,0.25);
        \node at (14.8,-0.35) {\scriptsize $\mathsf{Q_C}$} ;

        \draw[fill=modelcolor,opacity=1, rounded corners=1pt, thick] (14.35,2.25) rectangle (15.25,2.75); 
        \node at (14.8,2.5) {\scriptsize \textsf{\ctd}} ;
        \draw[   -latex, thick] (15.25,2.5) -- (15.75,2.5);
        \draw[fill=transgray,opacity=0.4, rounded corners=1pt,thick] (14.5,1.4) rectangle (15.1,1.9);
        \draw[  -latex, rounded corners=1pt, very thick] (14.8,1.9) -- (14.8,2.25);
        \node at (14.8,1.65) {\scriptsize $\mathsf{Q_2}$} ;

        \draw[  fill=modelcolor,opacity=1, rounded corners=1pt, thick] (14.35,4.25) rectangle (15.25,4.75); 
        \node at (14.8,4.5) {\scriptsize \textsf{\ctd}} ;
        \draw[   -latex, thick] (15.25,4.5) -- (15.75,4.5);
        \draw[  fill=transgray,opacity=0.4, rounded corners=1pt,thick] (14.5,3.4) rectangle (15.1,3.9);
        \draw[  -latex, rounded corners=1pt, very thick] (14.8,3.9) -- (14.8,4.25);
        \node at (14.8,3.65) {\scriptsize $\mathsf{Q_1}$} ;

        \draw[ black,thick] (13.75+2, 0.0) rectangle (15.25+2, 1.0);
        \draw[  dotted, rounded corners=1pt] (14.2+2,0.0) -- (14.2+2,1.0) ;
        \draw[thick, softblue, smooth] plot coordinates {
                (13.75+2, 0.45) (2.5 +11.5+2, 0.35)  (2.81 + 11.5+2, 0.35) (3.14 +11.5+2, 0.45) (3.47 +11.5+2, 0.3)
        };
        \foreach \x/\y in {11.5 + 2.81+2/0.35} {
                \draw[fill=softblue] (\x,\y) circle (0.07cm);
        }

        \draw[ black, thick] (13.75+2, 2) rectangle (15.25+2, 3);
        \draw[dotted, rounded corners=1pt] (14.2+2,2) -- (14.2+2,3) ;
        \draw[thick, softred, smooth] plot coordinates {
            (13.75+2, 2.7) (2.48 + 11.5+2, 2.6) (2.81+ 11.5+2, 2.45) (3.14 + 11.5+2, 2.4) (3.47+11.5+2, 2.6)
    };

    \foreach \x/\y in {3.14 +11.5+2 /2.4} {
            \draw[fill=softred] (\x,\y) circle (0.07cm);
    }

        \draw[black,thick] (13.75+2, 4.0) rectangle (15.25+2, 5.0);
          \draw[  dotted] (14.2+2,4.0) -- (14.2+2,5.0) ;
          \draw[thick, softgreen, smooth] plot coordinates {
            (13.75+2, 4.6) (2.48+11.5+2, 4.5) (2.81+11.5+2, 4.4) (3.14 + 11.5+2, 4.3) (3.47 +11.5+2, 4.25)
        };
        \foreach \x/\y in {2.95+11.5+2/4.36, 2.81+11.5+2/4.4, 3.47+11.5+2/4.25} {
            \draw[fill=softgreen] (\x,\y) circle (0.07cm);
        }


    \end{tikzpicture}
    }
    \caption{ Overview of \model{}'s architecture}\label{fig:model}
\end{figure*}

\section{Experiments}

We evaluate \model{} on four datasets: PhysioNet~\cite{Silva2012.Predicting}, MIMIC~\cite{Johnson2016.MIMICIII},
(Human) Activity~\cite{VedranaVidulin2010.Localization}, and USHCN~\cite{Menne2006.US}.
The baseline models already established in this setup originate from regular multivariate time series forecasting, 
IMTS classification and IMTS forecasting~\cite{Zhang.Irregular}.
DLinear~\cite{Zeng2023.Are}, TimesNet~\cite{Wu2022.TimesNet}, PatchTST~\cite{Nie2022.Time} and Crossformer~\cite{Zhang2022.Crossformer}
are well known models for regular multivariate forecasting.  
In a previous work~\cite{Zhang.Irregular} these RMTS forecasting models have been adapted to IMTS forecasting by treating timestamps and missingness indicators as additional input channels.
Following this approach, we additionally evaluate TSMixer~\cite{Chen2023.TSMixer}.

GRU-D~\cite{Che2018.Recurrent}
and mTAND~\cite{Shukla2020.MultiTime} are 
all models primarily introduced for IMTS classification, which are equipped with a predictor network which inputs 
hidden  IMTS representations and query times.  
Finally, Latent-ODE~\cite{Rubanova2019.Latent}, Continuous Recurrent Units (CRU)~\cite{Schirmer2022.Modeling}, Neural Flow~\cite{Bilos2021.Neural}, tPatchGNN~\cite{Zhang.Irregular} and GraFITi~\cite{Yalavarthi2023.Forecasting} 
are models designed for IMTS Forecasting.
To these baselines we add our own model and TimeCHEAT, a recently published IMTS forecasting model.
For TimeCHEAT, we use the implementations and hyperparameter recommendations as published in the paper introducing the model~\cite{Liu2025.Timecheat}.

To train \model{}, we implement the schedule-free AdamW\cite{Kingma2017.Adam,Loshchilov2019.Decoupled,Defazio2024.Road} optimizer with an initial learning rate of 0.01 and use early stopping with patience of 20 epochs along with a batch-size of 32. 
The hyperparameter tuning is conducted by selecting the best out of 20 randomly sampled configurations in terms of validation MSE.
We tune the aggregated-channel dimension $D\in\{64,128,256\}$, the output dimension $D_{\textsc{out}}\in\{32,64,128\}$, the number of mixer blocks $\in\{1,2,3\}$, and the weight decay $\in\{10^{-2},10^{-3},10^{-4}\}$.
\paragraph{Results.} 
\begin{table}[ht]
    \setlength{\tabcolsep}{2mm}
    \caption{Comparison of different algorithms across four datasets based on Test MSE.\@ 
            We reproduced the results from GraFITi~\cite{Yalavarthi2023.Forecasting} with the hyperparameters from its official GitHub repository.
            The results for \model{}, TSMixer and TimeCHEAT are from our own experiments.
            Other results are reported from the paper introducing tPatchGNN~\cite{Zhang.Irregular}.
            Each training run was repeated 5 times with different random seeds and the standard deviation in between these trials is given after the ±.
            We report the best results in \textbf{bold} and the second-best results \UL{underlined}.
            }\label{tab:main} 
    \centering
        \begin{tabular}{c l c c c c}
            \toprule
             & & PhysioNet & MIMIC & Activity & USHCN \\ 
            \midrule
             & & MSE$\times 10^{-3}$ & MSE$\times 10^{-2}$ & MSE$\times 10^{-3}$ & MSE$\times 10^{-1}$ \\ 
            \midrule
            \multirow{5}{*}{\scriptsize\shortstack{RMTS \\ Forc.}} & TSMixer & 8.66 ± 0.47 & 1.89 ± 0.01 & 3.01 ± 0.01 & 5.39 ± 0.04 \\
            & DLinear  & 41.86 ± 0.05 &  4.90 ± 0.00  & 4.03 ± 0.01 &  6.21 ± 0.00\\ 
            & TimesNet  & 16.48 ± 0.11 &  5.88 ± 0.08 & 3.12 ± 0.01 & 5.58 ± 0.05 \\ 
            & PatchTST  & 12.00 ± 0.23 &  3.78 ± 0.03 &  4.29 ± 0.14 & 5.75 ± 0.01 \\ 
            & Crossformer & 6.66 ± 0.11 &  2.65 ± 0.10 &  4.29 ± 0.20 & 5.25 ± 0.04 \\
            \midrule
            \multirow{2}{*}{\scriptsize \shortstack{IMTS \\ Class.}} & mTAND  & 6.23 ± 0.24 & 1.85 ± 0.06 & 3.22 ± 0.07 & 5.33 ± 0.05 \\ 
            & Latent-ODE  & 6.05 ± 0.57 & 1.89 ± 0.19 & 3.34 ± 0.11 & 5.62 ± 0.03 \\ 
            \midrule
            \multirow{6}{*}{\scriptsize\shortstack{IMTS \\ Forc.}} & Neural Flow  & 7.20 ± 0.07 & 1.87 ± 0.05 & 4.05 ± 0.13 & 5.35 ± 0.05 \\
            & CRU  & 8.56 ± 0.26 & 1.97 ± 0.02 & 6.97 ± 0.78 & 6.09 ± 0.17 \\ 
            & GraFITi & \UL{4.89 ± 0.12} & \BF{1.53 ± 0.02} & \UL{2.65 ± 0.02} & {5.17 ± 0.11} \\ 
            & TimeCHEAT & 6.18 ± 0.25 & 1.76 ± 0.01 & 4.42 ± 0.60 & 5.70 ± 0.53 \\
            & tPatchGNN  & {4.98 ± 0.08} & 1.69 ± 0.03 & 2.66 ± 0.03 & \UL{5.00 ± 0.04} \\
            & \textbf{\model{}} & \BF{4.88 ± 0.03} & \UL{1.61 ± 0.01} & \BF{2.50 ± 0.01} & \BF{4.91  ± 0.05} \\ 
            \bottomrule
        \end{tabular}
\end{table}
\Cref{tab:main} reports the forecasting accuracy of \model{} compared to the baseline models.
\model{} provides the most accurate forecasts on PhysioNet, Activity and USHCN.\@ For MIMIC, our model comes in second, closely behind GraFITi. 
We observe that GraFITi's relative performance increases with the number of channels. 
While it outperforms \model{} on MIMIC (102 chan.) and is on par with our model on PhysioNet (37 chan.), it is significantly outperformed by \model{} on Activity (12 chan.) and USHCN (5 chan.).

Additionally, we observe that TimeCHEAT does not provide competitive forecasting accuracy despite being the most recent baseline (AAAI 2025). 
The model is likely struggling, due to its inability to account for different query time steps. 
This major flaw is less problematic in very short-term forecasting scenarios, as used in the paper that introduced TimeCHEAT~\cite{Liu2025.Timecheat}. 
\paragraph{Efficiency Analysis.}
We conduct our experiments on an NVIDIA~1080Ti with 12~GB of GPU Memory.
In \Cref{tab:effi}, we compare the inference time and parameter count of recent models to that of \model{}.
Here, inference time refers to the total time it takes to answer all queries from the test set using a batch size of 32.
We observe that \model{}'s inference time is consistently shorter than the inference time of competing models.
Furthermore, it uses the fewest number of parameters on three out of four datasets.
\begin{table}[]
    \centering
    \small
    \setlength{\tabcolsep}{2mm}
    \caption{Number of parameters (left) and inference time (right) in seconds for each dataset}\label{tab:effi}
    \begin{tabular}{l | cc |cc| cc| cc}
        \toprule
        & \multicolumn{2}{c}{PhysioNet} & \multicolumn{2}{c}{MIMIC} & \multicolumn{2}{c}{Activity} & \multicolumn{2}{c}{USHCN} \\ 
        \midrule
        & \#P & t\_inf  &\#P & t\_inf  & \#P & t\_inf  & \#P & t\_inf  \\ 
        \midrule     
        GraFITi  & 251k & 1.9s  &  \textbf{255k} & 4.0s & 249k  & 0.6s & 496k & 2.2s \\
        tPatchGNN & 362k & 2.7s & 363k & 10s& 166k & 1.4s & 167k & 2.5s \\
        TimeCHEAT & 659k & 33s & 755k & 72s   & 1262k & 15s &  330k & 63s \\
        \model{} & \BF{207k} &\BF{0.7s} & {497k} & \BF{2.9s} & \BF{72k} & \BF{0.3s} & \BF{40k}& \BF{1.4s} \\
        \bottomrule
    \end{tabular}
\end{table}

\section{Limitations}
Theoretically, \model{} is limited by its fixed-size channel aggregation, which converts the sequence of observations from a channel into a fixed-sized embedding.
For very long sequences, or when typical sequence lengths vary significantly across channels, the fixed-size representation may become a bottleneck or fail to adequately capture the diversity across channels.
However, in practice, we do not find this to be an issue.
Our model is able to fit all the common benchmark datasets well with varying number of channels and distribution of sequence lengths.
Another limitation is that the parameter count in the mixer blocks scales quadratically with the number of channels. 
We observe that \model{} struggles with datasets that contain many channels. 
\model{} is outperformed by GraFITi only on the MIMIC dataset, which has the most channels and is on par with GraFITi on PhysioNet, which has the second-most channels.
Furthermore, we find that MIMIC is the only dataset in which IMTS-Mixer does not have the lowest number of parameters, as shown in \Cref{tab:effi}.
However, on datasets with fewer channels (e.g., Activity, USHCN), our model achieves significantly higher forecasting accuracy than competing methods.

\section{Conclusion and Future Work}
In this work we present \model{}, a parameter efficient and all MLP-based model for IMTS forecasting.
Our architecture incorporates ISCAM, a novel method for aggregating irregularly sampled univariate time series that, despite its simplicity, outperforms existing approaches.
Additionally, we generalize linear temporal projection layers to IMTS forecasting targets by introducing \ctd{} and yield equal or better performance than previously applied solutions. 
In our experiments, we show that \model{} establishes a new state-of-the-art forecasting accuracy on
three out of four evaluated datasets, despite its architectural simplicity.
We note that our model has difficulty handling time series with a large number of channels, as the parameter count in the fully connected layers increases significantly, which is associated with higher test error.
For future work, we plan on expanding our model to the related fields of IMTS classification and interpolation.
Furthermore, we will investigate \model{}'s applicability to probabilistic IMTS forecasting, where it could serve as an encoder of conditional normalizing flows~\cite{Yalavarthi2025.Probabilistic,Yalavarthi2025.Reliable}.

\bibliographystyle{splncs04}
\bibliography{./references}

\appendix
\section{Datasets}\label{app:dataset}    

\textbf{PhysioNet}~\cite{Silva2012.Predicting} and \textbf{MIMIC}~\cite{Johnson2016.MIMICIII} both contain the vital signs of intensive care unit patients. 
To evaluate the models, we query them to forecast all the measured variables for 24 hours based on all observations from the initial 24 hours after a patient's admission.

\textbf{Human Activity}~\cite{VedranaVidulin2010.Localization} contains 3-dimensional positional records of four sensors attached resulting in 12 variables. 
The sensors are attached to individuals who perform various activities (walking, sitting among others). 
Models are tasked with predicting 1 second of human motion based on 3 seconds of observation.

Finally, \textbf{USHCN}~\cite{Menne2006.US} is created by combining daily meteorological measurements from over one thousand weather stations that are distributed over the US.\@
While the 3 datasets mentioned above contain sparsely sampled IMTS intrinsically, USHCN originally was observed regularly and transformed into IMTS by randomly sampling observations. 
In each IMTS we use 2 years of observations to predict the climate conditions of a single month.
For each dataset we split the samples into 60\% train 20\% validation and 20\% test data.
Additional information about the datasets are given in \Cref{tab:dataset_summary}.
We calculate the Sparsity of each dataset by dividing the number of non-missing observations ($N_\textsc{obs}$) by the 
number of given observation steps ($N_T$) times the number of channels ($C$): 
$\text{Sparsity} = \frac{N_\textsc{obs}}{N_T C }$

Previous works~\cite{DeBrouwer2019.GRUODEBayes,Bilos2021.Neural,Schirmer2022.Modeling,Yalavarthi2023.Forecasting} used the parts of these datasets, but with different preprocessing, chunking and validation protocols.
We want to emphasize that therefore, the results reported in these works are incomparable.

\begin{table}[h]
    \small
    \setlength{\tabcolsep}{1mm}
    \caption{Summary of dataset characteristics. 
            \emph{Obs. / Forc. Range} refers to the observation range and forecasting horizon.
            \emph{Instances} are the IMTS from training- validation- and test set combined.
            \emph{Channels} are the number of variables observed.
            \emph{min/max avg. Obs.} refers to the minimum/maximum number of observation per channel averaged over instances.}
    \centering
    \begin{tabular}{lcccc}
    \toprule
    & {PhysioNet} & {MIMIC} & {H. Activity} & {USHCN} \\ 
    \midrule
    Obs. / Forc.~range & 24h/24h & 24h/24h & 3s/1s & 2y/1m\\
    Sparsity & 86\%    & 97\%    & 75\%  & 78\% \\
    Instances & 12,000 & 23,457 & 5,400 & 26,736 \\ 
    Channels & 41                 & 96             & 12   & 5 \\ 
    min avg. Obs. & 0.07               &  0.0007            & 19.7                   & 32.7           \\ 
    max avg. Obs. & 31.1               & 3.3            & 23.9                   & 35.6           \\ 
    
    \bottomrule
    \end{tabular}\label{tab:dataset_summary}
\end{table} 

\section{Experimental Details} \label{app:exp}
\paragraph{\model{}}
For the PMAU benchmark, we sample 20 hyperparameter configurations and select the one with the lowest MSE on the validation split.
For Physiome-ODE, we adhere to the proposed protocol and conduct random search with 10 configurations and evaluating them on the validation set of a single fold~\cite{Klotergens2024.PhysiomeODE}.
Furthermore, we set the hidden dimension of the non-linear networks that encode observation and query time stamps to 32.
To train \model{}, we implement the schedule-free~\cite{Defazio2024.Road} AdamW\cite{Kingma2017.Adam,Loshchilov2019.Decoupled} optimizer with an initial learning rate of 0.01 and use early stopping with patience of 20 epochs.
We use a batch-size of 32. 
\begin{itemize}
    \item We tune the dimension of the aggregated channels $D$ from the set \{64,128,256 \}
    \item The range for $D_\textsc{out}$ is \{32,64,128\}
    \item We allow the number of mixer blocks to be   \{1,2,3\}
    \item We tune the weight decay from \{1e-2, 1e-3, 1e-4\}
\end{itemize}

\paragraph{GraFITi}\label{app:graf}
We refer to the hyperparameters as given in the GitHub repository from GraFITi~\cite{Yalavarthi2023.Forecasting}:
\newline
\url{https://github.com/yalavarthivk/GraFITi}
\begin{itemize}
    \item PhysioNet: 4 layers, 4 heads, latent-dimension of 64 and a batch size of 32
    \item MIMIC: 4 layers, 4 heads, latent-dimension of 64 and a batch size of 32
    \item Activity: 4 layers, 4 heads, latent-dimension of 64 and a batch size of 32
    \item USHCN: 2 layers, 2 heads, latent-dimension of 128 and a batch size of 32
\end{itemize}
To compare our model with GraFITi on the Physiome-ODE benchmark in terms of parameter count and inference term,
we retain the hyperparameter parameters by following the search protocol as described in~\cite{Klotergens2024.PhysiomeODE}.

\paragraph{TimeCHEAT}~\cite{Liu2025.Timecheat}
We used the code as published in: \newline \url{https://github.com/Alrash/TimeCHEAT}.
The batch-size is set to 32 and following the published implementation, we use 
a decay on plateau scheduler with Adam optimizer and an initial learning of 0.001.
For TimeCHEAT we tuned the hyperparameters from the following grid:
\begin{itemize}
    \item Number of transformer layers: \{1, 2, 3\}
    \item Transformer hidden dimension: \{64, 128, 256 \}
    \item Number of encoder layers: \{1, 2, 4\}
    \item The initial learning rate is tuned from  \{1e-4, 1e-3\}
\end{itemize}

\paragraph{TSMixer}
We use PyTorch implementation from TS-Mixer~\cite{Chen2023.TSMixer} as published here: \newline
\url{https://github.com/ditschuk/pytorch-tsmixer}.
TSMixer is applied as follows:
Timestamps and a binary mask indicating whether a channel is observed at a given time step are used as additional variables, which are not forecasted.
Query timestamps are input into the model as future covariates.
The model is trained using the AdamW optimizer, with a weight decay of 0.0001 and a batch size of 32.
We tune these following hyperparameters by random search from the these ranges:
\begin{itemize}
    \item The initial learning rate is tuned from  \{1e-4, 1e-3\}
    \item Dropout: \{0.0, 0.1, 0.2 \}
    \item Number of Mixer Blocks: \{1, 2, 3\}
\end{itemize}

\paragraph{tPatchGNN}\label{app:tpatch}
For tPatchGNN~\cite{Zhang.Irregular}, we use the implementation as published here:
\newline
\url{https://github.com/usail-hkust/t-PatchGNN}
\begin{itemize}
    \item The hidden dimension is selected from $\{32,{64},128\}$
    \item The time embedding dimension is taken from  $\{5,{10},20\}$
    \item The initial learning rate is tuned from  \{1e-4, 1e-3\}
    \item The patch size
    \begin{itemize}
        \item is tuned from $\{{2},4,{8}\}$ for PhysioNet, MIMIC, USHCN
        \item is tuned from $\{100,{300},750\}$ for Activity
    \end{itemize} 
\end{itemize}

\section{Physiome-ODE}~\label{app:physiome}
\begin{table}[]
\centering
\caption{We report the average rank based on Test MSE, assigning the better rank in case of ties. \# Wins indicates the number of datasets where a model achieves the lowest Test MSE.
\%Gap-MSE refers to the mean relative difference in test MSE compared to the best model on each dataset.}
\begin{tabular}{lccc}
\toprule
 & Avg. Rank & \# Wins & \%Gap-MSE \\
\midrule
GraFITi-C      & 4.70 & 6  & 135 \\
Neural Flows   & 7.68 & 0  & 343 \\
CRU            & 4.44 & 2  & 74.4 \\
LinODENet      & 3.36 & \UL{7}  & \UL{21.1} \\
GraFITi        & \UL{3.26} & 2  & 41.6 \\
tPatchGNN      & 4.78 & 1  & 79 \\
TimeCHEAT      & 5.80 & 1  & 141 \\
IMTS-Mixer     &  \textbf{1.20} & \textbf{42} & \textbf{4.1} \\
\bottomrule
\end{tabular}

\end{table}
In order to provide a broader evaluation we compare \model{} on the Physiome-ODE benchmark.
The 50 datasets, that are contained in Physiome-ODE were created by solving Biological ODEs for 100 steps.
80\% of the resulting \emph{observations} are dropped uniform at random.
Each dataset consists of 2000 IMTS that each have different ODE constants and initial states. 
We refer to  PhysiomeODE~\cite{Klotergens2024.PhysiomeODE} for more details.
\model{} is compared with GraFITi, LinODENet~\cite{Scholz2022.Latent}, CRU, Neural Flows and GraFITi-C, a constant variant of GraFITi.
Additionally, we add tPatchGNN and TimeCHEAT to the benchmark.

The Physiome-ODE benchmark was originally proposed to highlight the strengths of neural ODEs~\cite{Bilos2021.Neural,Schirmer2022.Modeling}
and in first experiments a member of that model family (LinODENet) had the best forecasting accuracy on most datasets.
Now, \model{} has the lowest test MSE on 42 of 50 datasets, establishing a new state-of-the-art on this benchmark.
We list the test MSEs for all 50 datasets in \Cref{app:physiome}.

We report the test MSE of GraFITi-C~\cite{Klotergens2024.PhysiomeODE}, Neural Flows~\cite{Bilos2021.Neural}, 
CRU~\cite{Schirmer2022.Modeling}, LinODENet\cite{Scholz2022.Latent}, GraFITi~\cite{Yalavarthi2023.Forecasting},
tPatchGNN~\cite{Zhang.Irregular}, TimeCHEAT~\cite{Liu2025.Timecheat}, and \model{} on each dataset of the  Physiome-ODE
benchmark~\cite{Klotergens2024.PhysiomeODE} in \Cref{tab:pode}.

The number of parameters from each model for each dataset is shown in \Cref{tab:podeparams}.

Additionally, we list the respective inference times in \Cref{tab:podeinftime}.
For all models, we use an NVIDIA 1080Ti and a batch size of 32.

\section{Evaluating Alternative Components}~\label{sec:abl}
We conduct an ablation study in which we replace ISCAM, with Multi-Head Attention (MHA)~\cite{Vaswani2017.Attention} and Transformable Time-aware Convolution Network (TTCN)~\cite{Zhang2024.Irregulara}.
When we replace ISCAM with MHA, we infer the hidden state of channel $c$ with:
\begin{align*}
    Z_c^{\textsc{mha}} &= \mathcal{H}_c + \text{ReLU} \Bigl( \text{Linear}_\textsc{mha}(\mathcal{H}) \Bigr) , \quad \text{with} \\
    \mathcal{H}_c &= \text{MHA} \Bigl(\mathcal{Q}_c, \mathcal{K}_c, \mathcal{V}_c \Bigr) 
\end{align*}
To aggregate a channel into a fixed-sized vector, we use a singular query $\mathcal{Q}_c$ that is implemented with a learnable channel representation.
Following previous work~\cite{Yalavarthi2023.Forecasting,Liu2025.Timecheat}, we use the sequence of encoded time-value tuples as keys and values ($\mathcal{K}_c = \mathcal{V}_c$).
Here, these tuples are encoded by concatenating the scalar value to a sinusoidal time-embedding.

In the hyperparameter tuning of this ablation study we additionally select the number of attention heads from \newline \{1, 2, 4, 8\}.The result is shown in \Cref{tab:enc}. For a fair comparison, we re-tune the hyperparameters for each \model{} modification.
Our results show, that among the evaluated candidates ISCAM is the best encoder for our model. 

Furthermore, we compare \ctd{} with using an MLP as temporal projection as implemented in tPatchGNN. 
Here, a sinusoidal embedding of the query times is concatenated to the hidden state of each channel
and the forecast is obtained by a 2-layer MLP that inputs the enriched channel representation and outputs a scalar~\cite{Zhang.Irregular},   
On each dataset \ctd{} shows to be more or equally effective than the existing MLP based method.

\begin{table}[]
    \centering
    \caption{Comparing Test MSE of \model{} variants with different Encoders and Temporal Projection modules.}\label{tab:enc}
    \begin{tabular}{l c c c c}
        \toprule
         & PhysioNet & MIMIC & H. Activity & USHCN \\ 
        \midrule
        MSE& {$\times 10^{-3}$} & {$\times 10^{-2}$} & { $\times 10^{-3}$} & { $\times 10^{-1}$} \\ 
        \midrule
        & \multicolumn{4}{c}{Encoder} \\
        TTCN & 5.07 ± 0.04 & 1.71 ± 0.05 & 2.54 ± 0.02      & \UL{4.94 ± 0.06} \\
        MHA &  \UL{4.94 ± 0.06} & \UL{1.66 ± 0.01} &  \UL{2.52 ± 0.01} & \UL{4.94 ± 0.05}  \\
        ISCAM & \BF{4.88 ± 0.03} & \BF{1.61 ± 0.01} & \BF{2.50 ± 0.01} & \BF{4.91 ± 0.05}\\
        \midrule
        & \multicolumn{4}{c}{Temporal Projection} \\
        MLP  & {4.91 ± 0.03} & \BF{1.61 ± 0.01} &   2.51 ± 0.02 & 5.11 ± 0.03 \\
        \ctd & \BF{4.88 ± 0.03} &  \BF{1.61 ± 0.01} & \BF{2.50 ± 0.01} & \BF{4.91 ± 0.05}  \\ 
        \bottomrule
    \end{tabular}
\end{table}

\subsection{Effect of Mixer Blocks}
\begin{figure}[h!]
    \centering
    \begin{tabular}{cc}
      \includegraphics[width=0.35\textwidth]{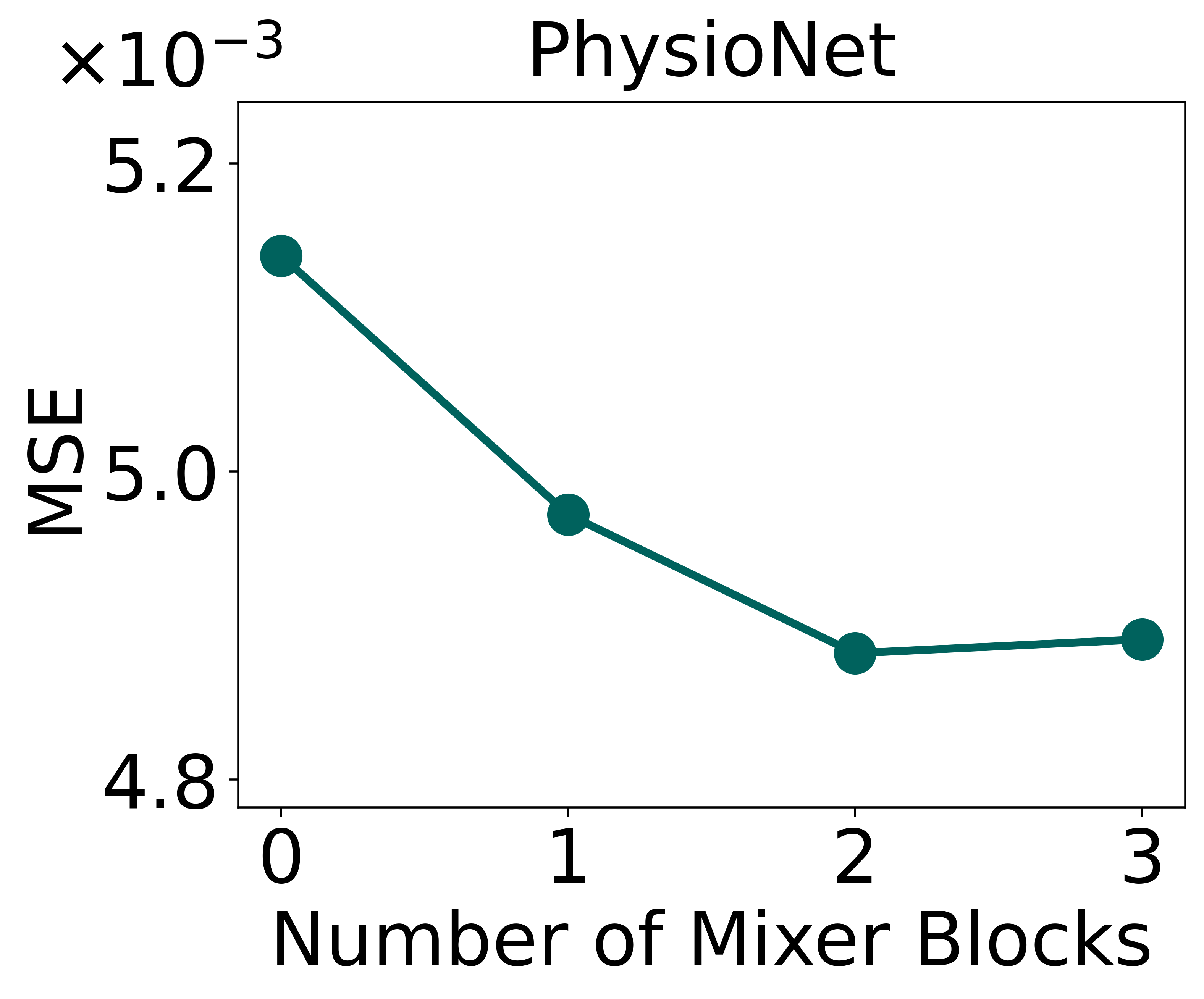} & 
      \includegraphics[width=0.35\textwidth]{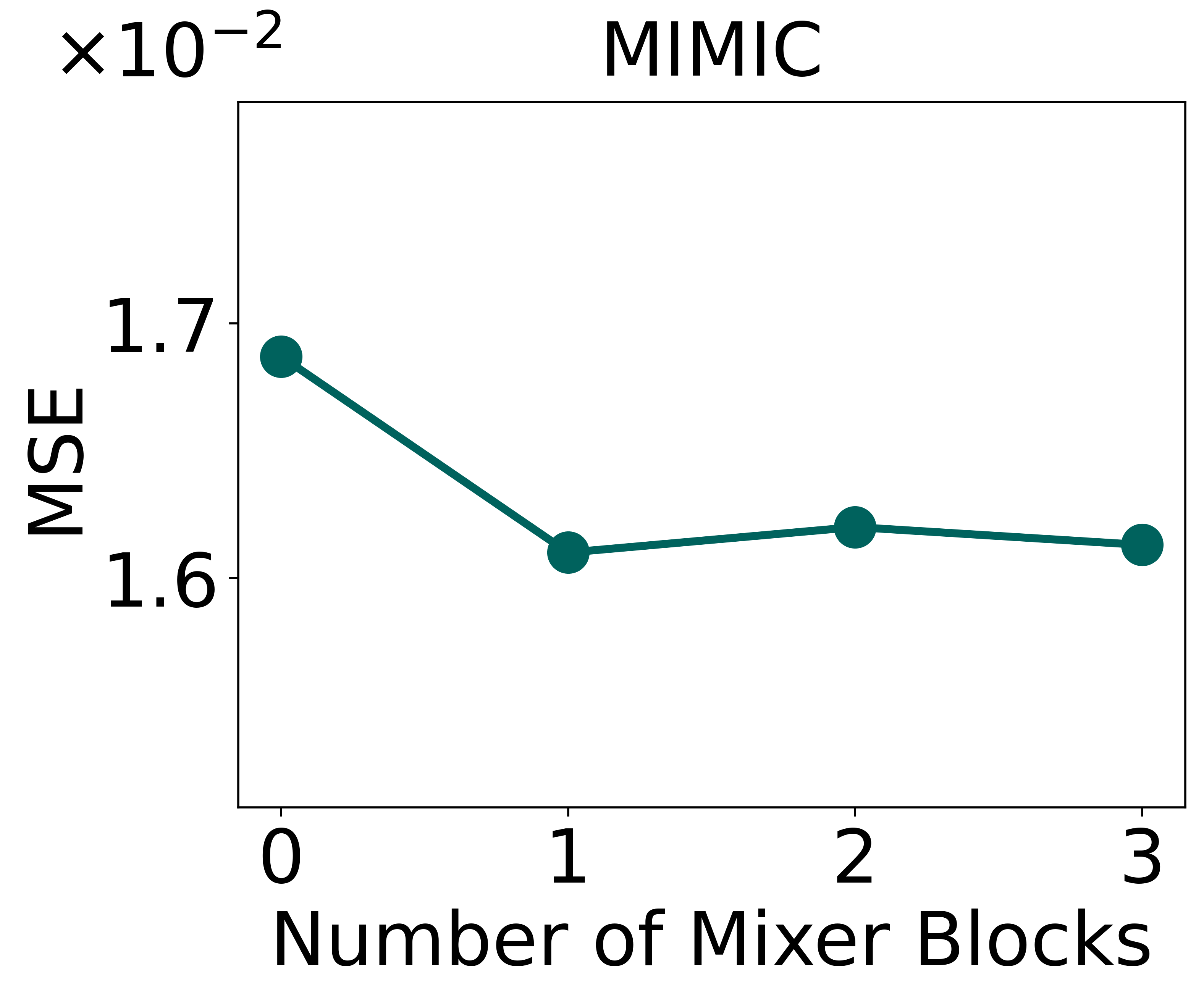} \\
      \includegraphics[width=0.35\textwidth]{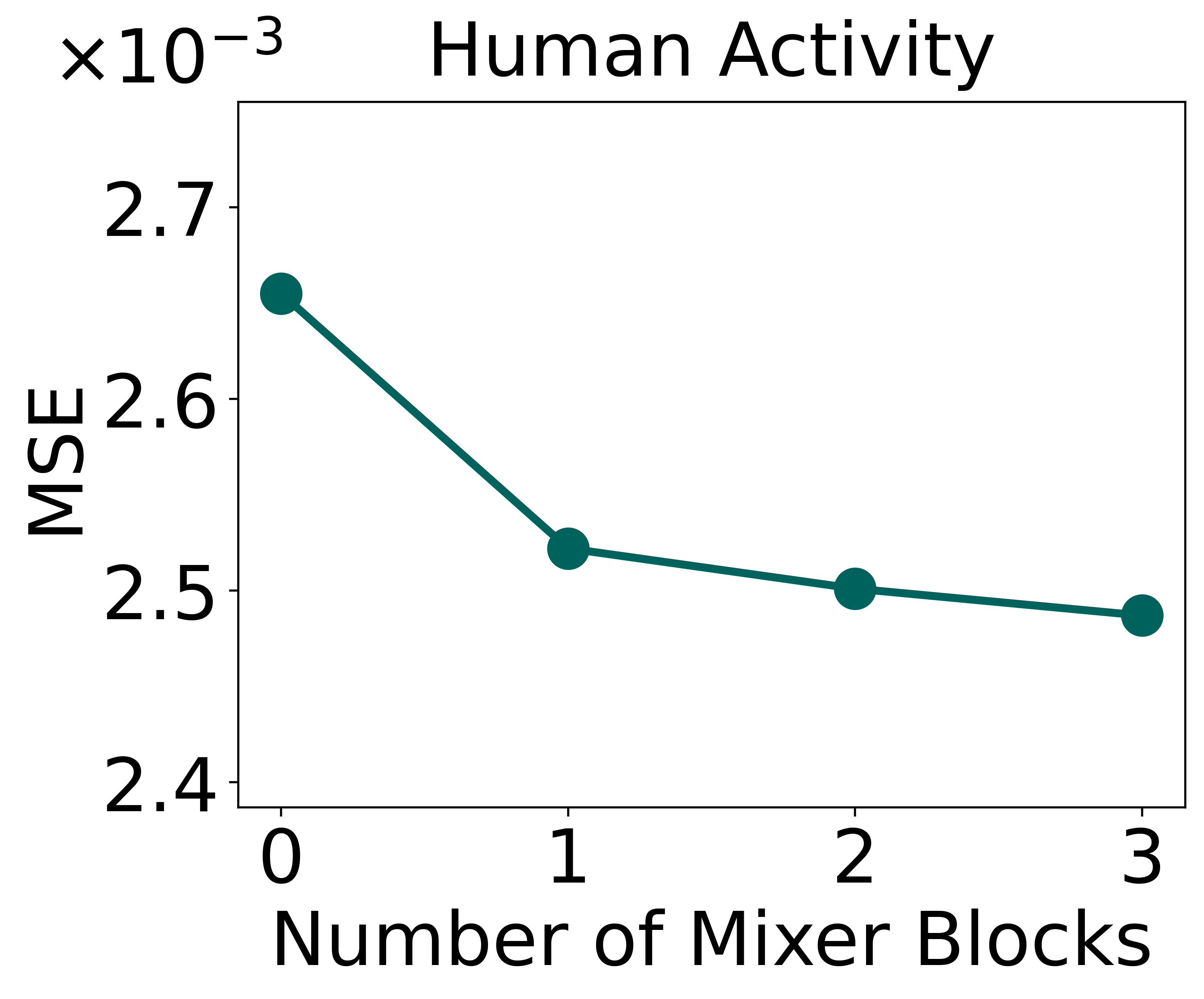} &
      \includegraphics[width=0.35\textwidth]{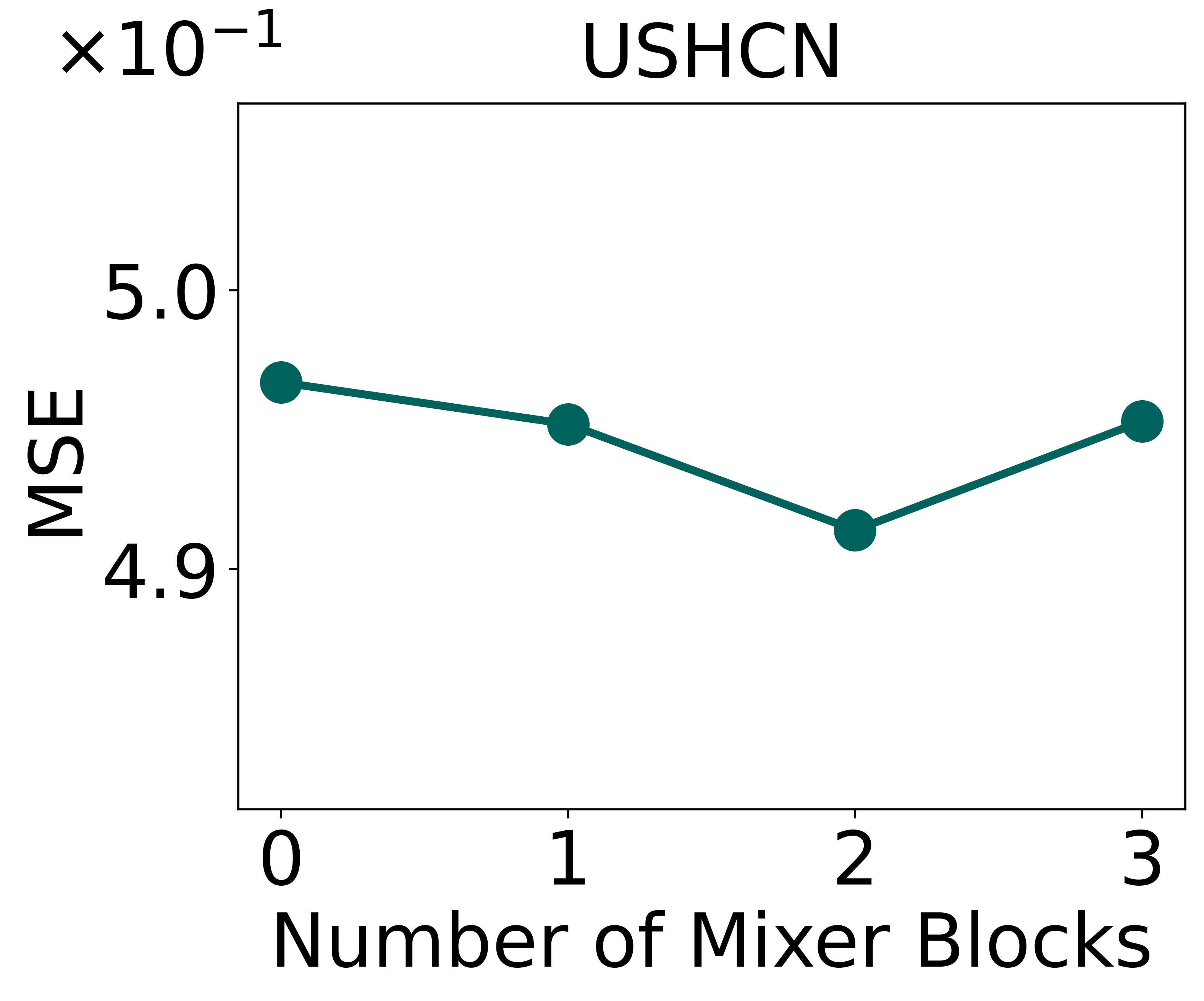}
  \end{tabular}
  \caption{Test MSE with varying number of mixer blocks.}~\label{fig:mixblocks}
\end{figure}

We conducted an experiment to assess the effect of that the number of mixer layers has on the forecasting accuracy on the four PMAU datasets. 
The results shown in \Cref{fig:mixblocks} indicate that having at least one mixer block is necessary to obtain the optimal performance.
However, stacking more and more mixer blocks is usually not beneficial on the evaluated datasets. 
Furthermore, we observe that \model{}, without a single mixer block, is already competitive with state-of-the-art baselines, showcasing the
ability of ISCAM and \ctd.

\begin{table*}
  \caption{Test MSE on the Physiome-ODE~\cite{Klotergens2024.PhysiomeODE} benchmark. The results for the competing models are reported from the Physiome-ODE paper. }\label{tab:pode}
\scriptsize
\setlength{\tabcolsep}{1mm}
\centering
\begin{tabular}{lcccccccc}
\toprule
 & \multirow{2}{*}{IMTS-} & \multirow{2}{*}{Time-} & \multirow{2}{*}{tPatch-} & \multirow{2}{*}{GraFITi} & \multirow{2}{*}{LinODE-} & & \multirow{2}{*}{} \\
 Dataset & \multirow{2}{*}{Mixer} & \multirow{2}{*}{CHEAT} & \multirow{2}{*}{GNN} & & \multirow{2}{*}{Net} & CRU & {GraFITi-C} \\ \\
\midrule
  DUP01 & \UL{0.954±.038} &      0.978±.030 &      0.957±.035 &      0.955±.037 &      0.964±.036 &      0.958±.038 & \BF{0.951±.036} \\
  JEL01 & \BF{0.935±.018} &      0.966±.027 &      0.943±.016 &      0.942±.020 &      0.949±.015 &      0.939±.015 & \BF{0.935±.016} \\
  DOK01 & \BF{0.978±.005} &      0.997±.004 &      0.987±.004 &      0.984±.005 &      0.996±.003 &      0.985±.005 & \UL{0.982±.005} \\
  INA01 & \BF{1.003±.009} &      1.005±.010 &  \UL{1.004±.009} & \UL{1.004±.010} &      1.009±.011 &     1.005±.010 & \UL{1.004±.009} \\
  WOL01 & \BF{0.783±.026} &      0.804±.029 &      0.795±.027 &      0.787±.028 &      0.806±.027 &      0.814±.029 & \UL{0.784±.030} \\
  BOR01 & \BF{0.708±.021} &      0.718±.022 &      0.719±.019 &      0.712±.022 &      0.719±.021 &      0.715±.020 & \UL{0.709±.022} \\
  HYN01 & \BF{0.608±.048} &      0.652±.049 &      0.630±.045 &      0.625±.043 &      0.672±.044 &      0.665±.053 & \UL{0.619±.046} \\
  JEL02 & \BF{0.664±.030} &      0.714±.027 &      0.702±.023 &      0.699±.029 &      0.693±.031 & \UL{0.674±.028} &     0.687±.027 \\
  DUP02 & \UL{0.719±.046} &      0.748±.047 &      0.728±.049 &      0.728±.044 &      0.740±.042 &      0.722±.046 & \BF{0.718±.046} \\
  WOL02 & \BF{0.640±.014} &      0.679±.012 &      0.656±.016 &      0.654±.014 &      0.663±.015 &      0.653±.017 & \UL{0.645±.016} \\
  DIF01 & \UL{0.959±.029} &      0.977±.031 &      0.991±.027 &      0.985±.030 & \BF{0.832±.087} &      0.985±.025 &     0.982±.029 \\
  VAN01 & \BF{0.240±.005} &      0.252±.007 &      0.247±.007 &      0.246±.005 &      0.250±.006 &      0.253±.005 & \UL{0.242±.006} \\
  DUP03 & \BF{0.610±.050} &      0.683±.046 &      0.624±.052 &      0.627±.043 &      0.632±.044 & \UL{0.622±.047} &     0.744±.042 \\
  BER01 & \BF{0.277±.013} &      0.311±.028 &      0.310±.014 &      0.300±.018 & \UL{0.279±.020} &      0.280±.016 &     0.342±.018 \\
  LEN01 & \UL{0.547±.043} &      0.932±.112 &      0.981±.062 &      0.607±.055 & \BF{0.387±.071} &      0.754±.157 &     0.970±.063 \\
   LI01 &      0.177±.018 &      0.438±.019 &      0.394±.028 &      0.202±.013 & \BF{0.084±.009} & \UL{0.175±.020} &     0.742±.010 \\
   LI02 & \BF{0.360±.047} &      0.464±.048 &      0.438±.048 & \UL{0.397±.058} &      0.434±.044 &      0.437±.046 &     0.458±.056 \\
  REV01 &      0.603±.045 &      0.741±.057 &      0.741±.054 &      0.674±.055 & \BF{0.597±.061} & \UL{0.602±.049} &     0.855±.050 \\
  PUR01 & \UL{0.143±.017} &      0.343±.017 &      0.397±.082 &      0.153±.006 & \BF{0.106±.006} &      0.353±.083 &     0.476±.020 \\
  NYG01 & \BF{0.237±.051} &      0.402±.061 &      0.357±.067 & \UL{0.344±.065} &      0.358±.071 &      0.403±.092 &     0.366±.047 \\
  PUR02 & \BF{0.251±.022} &      0.474±.057 &      0.390±.017 &      0.322±.021 & \UL{0.280±.028} &      0.293±.026 &     0.511±.023 \\
  HOD01 & \BF{0.383±.052} &      0.511±.043 &      0.479±.057 &      0.493±.046 &      0.441±.043 & \UL{0.409±.049} &     0.609±.056 \\
  REE01 & \BF{0.031±.005} &      0.044±.007 &      0.047±.011 & \UL{0.033±.007} &      0.045±.012 &      0.051±.008 &     0.039±.012 \\
  VIL01 & \BF{0.302±.038} &      0.385±.045 &      0.354±.028 & \UL{0.344±.044} &      0.374±.021 &      0.373±.039 &     0.378±.042 \\
  KAR01 & \BF{0.033±.012} &      0.050±.013 &      0.048±.012 &      0.041±.013 & \UL{0.034±.008} &      0.044±.012 &     0.078±.011 \\
  SHO01 & \BF{0.053±.007} &      0.085±.019 &      0.078±.012 &      0.062±.013 &      0.057±.006 &      0.095±.010 & \UL{0.055±.013} \\
  BUT01 & \BF{0.180±.036} &      0.306±.079 &      0.354±.102 &      0.281±.071 & \UL{0.254±.074} &      0.317±.108 &     0.324±.091 \\
  MAL01 & \BF{0.014±.004} &      0.044±.002 &      0.038±.011 &      0.020±.004 & \UL{0.018±.007} &      0.064±.007 &     0.054±.005 \\
  ASL01 & \BF{0.013±.004} &      0.029±.003 &      0.031±.004 &      0.025±.009 & \UL{0.022±.003} &      0.046±.014 &     0.026±.002 \\
  BUT02 & \BF{0.156±.022} &      0.272±.031 &      0.280±.043 &      0.248±.052 & \UL{0.207±.056} &      0.282±.042 &     0.256±.039 \\
  MIT01 & \BF{0.003±.000} &  \BF{0.003±  .000}& \BF{0.003±.000} & \BF{0.003±.000} & \BF{0.003±.000} &\BF{0.003±.000}& \BF{0.003±.000} \\
  GUP01 & \BF{0.011±.004} &      0.060±.014 &      0.047±.016 &      0.041±.006 & \UL{0.018±.007} &      0.057±.017 &     0.035±.006 \\
  GUY01 & \BF{0.004±.001} &      0.015±.004 &      0.005±.001 &      0.005±.003 &      0.006±.005 & \BF{0.004±.001} & \BF{0.004±.001} \\
  PHI01 & \BF{0.105±.022} &      0.308±.019 &      0.249±.032 &      0.222±.013 & \UL{0.131±.014} &      0.133±.020 &     0.345±.015 \\
  GUY02 & \BF{0.007±.002} &      0.032±.006 &      0.016±.003 &      0.012±.009 & \UL{0.010±.006} & \UL{0.010±.002} &     0.032±.015 \\
  PUL01 & \BF{0.006±.002} &      0.036±.011 &      0.018±.006 & \UL{0.008±.001} & \UL{0.008±.004} &      0.012±.003 &     0.024±.008 \\
  CAL01 & \BF{0.074±.005} &      0.720±.357 &      0.212±.033 &      0.179±.012 & \UL{0.078±.009} &      0.158±.008 &     0.643±.024 \\
  WOD01 & \BF{0.098±.007} &      0.372±.293 &      0.165±.012 &      0.164±.013 &      0.154±.016 & \UL{0.113±.017} &     0.344±.016 \\
  GUP02 & \BF{0.397±.016} &      0.445±.018 &      0.463±.024 &      0.449±.027 &      0.469±.022 & \UL{0.444±.018} &     0.461±.025 \\
    M01 & \BF{0.003±.000} &      0.013±.002 &      0.004±.001 & \BF{0.003±.000} &      0.004±.001 &      0.005±.000 & \BF{0.003±.000} \\
  LEN02 & \BF{0.031±.005} &      0.217±.108 &      0.089±.011 &      0.099±.021 & \UL{0.039±.005} &      0.059±.012 &     0.143±.022 \\
  KAR02 & \UL{0.143±.007} &      0.196±.008 &      0.160±.009 &      0.151±.009 & \BF{0.140±.010} &      0.151±.011 &     0.252±.010 \\
  SHO02 & \BF{0.021±.003} &      0.063±.015 &      0.070±.017 &      0.043±.006 & \UL{0.037±.006} &      0.083±.015 &     0.073±.010 \\
  MAC01 & \BF{0.016±.002} &      0.046±.005 &      0.033±.003 &      0.021±.003 &      0.020±.003 &      0.065±.006 & \UL{0.019±.002} \\
  IRI01 & \BF{0.027±.005} &      0.053±.011 &      0.048±.008 &      0.038±.017 & \UL{0.037±.003} &      0.049±.010 &     0.097±.008 \\
  BAG01 & \BF{0.026±.002} &      0.050±.003 &      0.045±.003 & \UL{0.029±.002} &      0.032±.005 &      0.046±.005 &     0.109±.002 \\
  WOL03 & \BF{0.062±.009} &      0.233±.053 &      0.142±.024 &      0.105±.016 & \UL{0.073±.010} &      0.177±.016 &     0.247±.032 \\
  WAN01 & \BF{0.082±.007} &      0.170±.014 &      0.140±.011 &      0.119±.010 & \UL{0.103±.012} &      0.125±.012 &     0.232±.015 \\
  NEL01 & \BF{0.006±.001} &      0.019±.001 &      0.013±.001 & \UL{0.007±.000} &      0.010±.001 &      0.009±.001 &     0.023±.006 \\
  HUA01 & \BF{0.037±.001} &      0.099±.006 &      0.090±.012 &      0.063±.005 & \UL{0.052±.004} &      0.116±.007 &     0.115±.007 \\
  \bottomrule
\end{tabular}
\end{table*}

\begin{table}[]
    \centering
    \small
    \setlength{\tabcolsep}{2mm}
    \caption{Exact number of trainable parameters on each Physiome-ODE dataset.}~\label{tab:podeparams}
    \begin{tabular}{lcccc}
    \toprule
    Dataset &  IMTS-Mixer &  TimeCHEAT &  tPatchGNN &  GraFITi \\
    \midrule
    DUP01 &      173,075 &    1,279,382 &     365,594 &   315,329 \\
    JEL01 &       39,053 &    1,279,382 &     390,170 &   236,673 \\
    DOK01 &      105,219 &     338,203 &     159,610 &  1,255,297 \\
    INA01 &      139,891 &     337,282 &     159,830 &    15,825 \\
    WOL01 &       95,195 &     344,723 &     390,310 &   315,777 \\
    BOR01 &      178,201 &     341,702 &     365,614 &  1,253,377 \\
    HYN01 &      315,631 &     339,608 &     365,994 &  1,255,809 \\
    JEL02 &       17,741 &     645,014 &     365,594 &   627,329 \\
    DUP02 &       42,317 &     341,702 &     165,454 &   627,457 \\
    WOL02 &      192,745 &     343,695 &     160,414 &   315,649 \\
    DIF01 &       52,321 &     659,850 &     159,570 &  1,255,041 \\
    VAN01 &       80,459 &     642,011 &     390,330 &  1,254,273 \\
    DUP03 &       56,083 &     653,849 &     365,594 &   158,017 \\
    BER01 &       95,625 &     641,920 &     365,774 &  1,254,401 \\
    LEN01 &       42,317 &     646,214 &     365,614 &   315,393 \\
    LI01  &       65,179 &     343,376 &     365,654 &   315,521 \\
    LI02  &       48,095 &     343,376 &     365,654 &  1,253,633 \\
    REV01 &       44,699 &    1,282,256 &     390,230 &  1,253,633 \\
    PUR01 &       59,109 &     341,702 &     165,454 &   315,393 \\
    NYG01 &      139,891 &     343,747 &     165,974 &  1,256,705 \\
    PUR02 &       65,165 &     646,214 &     165,454 &  1,253,377 \\
    HOD01 &       62,141 &     342,531 &     390,210 &  1,253,505 \\
    REE01 &       39,997 &     655,491 &     365,634 &   315,457 \\
    VIL01 &      236,175 &     334,849 &     165,574 &   315,777 \\
    KAR01 &      130,321 &     644,156 &     390,450 &  1,255,041 \\
    SHO01 &      700,145 &     679,820 &     166,514 &  1,260,161 \\
    BUT01 &       61,353 &     342,531 &     390,210 &  1,253,505 \\
    MAL01 &      236,725 &     344,336 &     165,574 &   315,777 \\
    ASL01 &      143,079 &     651,229 &     390,730 &  1,256,833 \\
    BUT02 &       62,141 &     342,531 &     390,210 &  1,253,505 \\
    MIT01 &      192,745 &     344,082 &     165,534 &   236,993 \\
    GUP01 &       39,997 &     332,915 &     165,474 &  1,253,505 \\
    GUY01 &      168,461 &     340,631 &     165,434 &     5,249 \\
    PHI01 &      199,845 &    1,280,582 &     159,310 &  1,253,377 \\
    GUY02 &      168,461 &     653,462 &     390,170 &   236,673 \\
    PUL01 &      168,461 &     653,462 &     159,290 &   158,017 \\
    CAL01 &      284,951 &    1,282,977 &     390,470 &  1,255,169 \\
    WOD01 &       56,473 &     638,710 &     165,454 &  1,253,377 \\
    GUP02 &       45,205 &     655,491 &     159,330 &  1,253,505 \\
      M01 &      193,811 &     340,502 &     365,594 &   158,017 \\
    LEN02 &      176,997 &     638,710 &     159,310 &  1,253,377 \\
    KAR02 &      115,255 &     335,936 &     390,390 &  1,254,657 \\
    SHO02 &      528,233 &     672,187 &     166,514 &  1,260,161 \\
    MAC01 &       52,913 &     343,953 &     390,270 &   315,649 \\
    IRI01 &       97,937 &     654,482 &     390,590 &   316,673 \\
    BAG01 &       85,813 &     657,425 &     365,734 &   628,225 \\
    WOL03 &      275,821 &     658,641 &     365,814 &  1,254,657 \\
    WAN01 &       44,699 &     640,642 &     159,350 &  1,253,633 \\
    NEL01 &       62,141 &     342,789 &     365,634 &  1,253,505 \\
    HUA01 &      130,093 &     348,077 &     365,914 &  1,255,297 \\
    \bottomrule
    \end{tabular}
\end{table}
\begin{table}[]
    \centering
    \small
    \setlength{\tabcolsep}{2mm}
    \caption{Inference time in seconds of each model on the datasets contained in the Physiome-ODE benchmark.}~\label{tab:podeinftime}
    \begin{tabular}{lcccc}
    \toprule
    Dataset &  IMTS-Mixer &  TimeCHEAT &  tPatchGNN &  GraFITi \\
    \midrule
     DUP01 &       0.064 &      2.772 &      0.088 &    1.508 \\
  JEL01 &       0.061 &      2.704 &      0.125 &    1.419 \\
  DOK01 &       0.078 &      2.377 &      0.117 &    1.596 \\
  INA01 &       0.087 &      2.511 &      0.136 &    1.503 \\
  WOL01 &       0.063 &      2.461 &      0.153 &    1.517 \\
  BOR01 &       0.059 &      2.509 &      0.095 &    1.504 \\
  HYN01 &       0.092 &      2.526 &      0.154 &    1.601 \\
  JEL02 &       0.058 &      2.617 &      0.090 &    1.343 \\
  DUP02 &       0.053 &      2.398 &      0.113 &    1.342 \\
  WOL02 &       0.066 &      2.520 &      0.110 &    1.482 \\
  DIF01 &       0.071 &      2.623 &      0.107 &    1.592 \\
  VAN01 &       0.063 &      2.360 &      0.148 &    1.404 \\
  DUP03 &       0.056 &      2.316 &      0.089 &    1.201 \\
  BER01 &       0.065 &      2.450 &      0.128 &    1.473 \\
  LEN01 &       0.054 &      2.555 &      0.093 &    1.482 \\
   LI01 &       0.064 &      2.383 &      0.103 &    1.504 \\
   LI02 &       0.058 &      2.484 &      0.100 &    1.514 \\
  REV01 &       0.061 &      2.708 &      0.129 &    1.505 \\
  PUR01 &       0.060 &      2.530 &      0.122 &    1.531 \\
  NYG01 &       0.097 &      2.479 &      0.222 &    1.598 \\
  PUR02 &       0.054 &      2.575 &      0.101 &    1.500 \\
  HOD01 &       0.061 &      2.500 &      0.129 &    1.530 \\
  REE01 &       0.062 &      2.640 &      0.096 &    1.505 \\
  VIL01 &       0.070 &      2.375 &      0.164 &    1.508 \\
  KAR01 &       0.073 &      2.484 &      0.224 &    1.536 \\
  SHO01 &       0.115 &      2.724 &      0.359 &    1.604 \\
  BUT01 &       0.058 &      2.376 &      0.117 &    1.493 \\
  MAL01 &       0.066 &      2.445 &      0.146 &    1.484 \\
  ASL01 &       0.076 &      2.313 &      0.300 &    1.387 \\
  BUT02 &       0.062 &      2.523 &      0.122 &    1.506 \\
  MIT01 &       0.062 &      2.344 &      0.126 &    1.290 \\
  GUP01 &       0.056 &      2.284 &      0.118 &    1.390 \\
  GUY01 &       0.063 &      2.432 &      0.116 &    1.265 \\
  PHI01 &       0.061 &      2.808 &      0.091 &    1.509 \\
  GUY02 &       0.060 &      2.599 &      0.121 &    1.442 \\
  PUL01 &       0.060 &      2.576 &      0.089 &    1.356 \\
  CAL01 &       0.075 &      2.537 &      0.213 &    1.462 \\
  WOD01 &       0.059 &      2.581 &      0.107 &    1.527 \\
  GUP02 &       0.055 &      2.644 &      0.089 &    1.502 \\
    M01 &       0.064 &      2.538 &      0.098 &    1.364 \\
  LEN02 &       0.061 &      2.566 &      0.095 &    1.523 \\
  KAR02 &       0.065 &      2.479 &      0.242 &    1.503 \\
  SHO02 &       0.141 &      3.060 &      0.286 &    1.711 \\
  MAC01 &       0.061 &      2.379 &      0.161 &    1.510 \\
  IRI01 &       0.072 &      2.504 &      0.197 &    1.482 \\
  BAG01 &       0.057 &      2.368 &      0.113 &    1.260 \\
  WOL03 &       0.068 &      2.498 &      0.137 &    1.483 \\
  WAN01 &       0.063 &      2.521 &      0.100 &    1.501 \\
  NEL01 &       0.056 &      2.315 &      0.093 &    1.366 \\
  HUA01 &       0.075 &      2.514 &      0.137 &    1.568 \\
    \bottomrule
    \end{tabular}
\end{table}

%
%

\end{document}